\documentclass[lettersize,journal]{IEEEtran}
\usepackage{amsmath,amsfonts}
\usepackage{amsthm}
\newtheorem{thm}{Rule}
\newtheorem{thm2}{Definition}
\usepackage{algorithm}
\usepackage[noend]{algpseudocode}
\algrenewcommand\textproc{}
\usepackage{array}
\usepackage[caption=false,font=normalsize,labelfont=sf,textfont=sf]{subfig}
\usepackage{textcomp}
\usepackage{stfloats}
\usepackage{url}
\usepackage{verbatim}
\usepackage{graphicx}
\usepackage{cite}
\usepackage{tabularx,booktabs}
\usepackage{multirow}

\begin{document}

\title{Toward Process Controlled Medical Robotic System}

\author{Yihao Liu, \IEEEmembership{Student Member, IEEE}, Amir Kheradmand, and Mehran Armand, \IEEEmembership{Senior Member, IEEE}
\thanks{Manuscript submitted July 17, 2023.} 
\thanks{This work was supported by grants from the National Institute of Deafness and Other Communication Disorders (R01DC018815), National Institute of Arthritis and Musculoskeletal and Skin Diseases (R01AR080315), and National Institute of Biomedical Imaging and Bioengineering (R01EB023939). }
\thanks{Yihao Liu is with the Biomechanical- and Image-Guided Surgical Systems Laboratory, Laboratory for Computational Sensing \& Robotics, and Department of Computer Science, Johns Hopkins University, Baltimore, MD 21218 USA (e-mail: yliu333@jhu.edu). }
\thanks{Amir Kheradmand is with the Department of Neurology, Department of Neuroscience, and Laboratory for Computational Sensing \& Robotics, Johns Hopkins University, Baltimore, MD 21218 USA.}
\thanks{Mehran Armand is with the Biomechanical- and Image-Guided Surgical Systems Laboratory, Laboratory for Computational Sensing \& Robotics, Department of Orthopaedic Surgery, Department of Mechanical Engineering, Department of Computer Science, and Applied Physics Laboratory, Johns Hopkins University, Baltimore, MD 21218 USA.}
}

\markboth{T\MakeLowercase{his work has been submitted to the} IEEE \MakeLowercase{for possible publication.} C\MakeLowercase{opyright may be transferred without notice, after which this version may no longer be accessible.}}
{Liu \MakeLowercase{\textit{et al.}}: Toward Process Controlled Medical Robotic System}


\maketitle

\begin{abstract}
Medical errors, defined as unintended acts either of omission or commission that cause the failure of medical actions, are the third leading cause of death in the United States. Medical errors can include communication breakdowns, diagnostic errors, poor judgment, and inadequate skills. The application of autonomy and robotics can alleviate some causes of medical errors by improving accuracy and providing means to accurately follow planned procedures. However, for the robotic applications to improve safety, they must maintain constant operating conditions in the presence of disturbances, and provide reliable measurements, evaluation, and control for each state of the procedure. This article addresses the need for process control in medical robotic systems, and proposes a standardized design cycle toward its automation. Monitoring and controlling the changing conditions in a medical or surgical environment necessitates a clear definition of workflows and their procedural dependencies. We propose integrating process control into medical robotic workflows using hierarchical Finite State Machines (hFSM) to identify changes in states of the system and environment, and execute possible operations or transitions to new states. Therefore, the system translates clinician experiences and procedure workflows into machine-interpretable languages. The design cycle using hFSM formulation can be a deterministic process, which opens up possibilities for higher-level automation in medical robotics. Shown in our work, with a standardized design cycle and software paradigm, we pave the way toward controlled workflows that can be automatically generated. Additionally, a modular design for a robotic system architecture that integrates hFSM can provide easy software and hardware integration. This article discusses the system design, software implementation, and example application to Robot-Assisted Transcranial Magnetic Stimulation (RATMS) and robot-assisted femoroplasty. We also provide assessments of these two system examples using their performance in terms of robotic tool placement and response to failure injections. 
\end{abstract}

\begin{IEEEkeywords}
medical robotics, surgical robotics, teleoperation, medical navigation system
\end{IEEEkeywords}

\section{Introduction}
\label{sec:introduction}

Medical robots can reduce procedural errors by performing tasks with accuracy and precision that may not be achievable by humans. Enhanced by mechanical accuracy and machine perception of the surroundings, robots can perform tasks that require a high level of dexterity and stability, while improving safety by defining forbidden zones and virtual fixtures \cite{rosenberg1993virtual, abbott2007haptic, bowyer2013active}. However, a gap still exists due to limited machine awareness of automated systems in medical procedures that involve dynamic and changing environments. Process control, aiming to monitor and adjust a process at the medical workflow level, can help medical robots effectively respond and adapt to the dynamics of the changing environments and improve human-robot interactions. The existing literature explores human-robot interactions and user experience but not within the context of a process-controlled medical workflow \cite{greatbatch1993interpersonal, azimi2018interactive, oppenheim2021mental, gao2021evaluation}. The workflows for medical robotics in current literature are primarily single-branched, which rely heavily on human decisions during transition steps. In these single-branched workflows, sometimes a failed step requires discarding and restarting the procedure all over again. 

Automated medical and surgical procedures use flags to indicate system status. However, software using excessive flags may become difficult to maintain. Software security issues may arise when flags are not explicitly updated. Complex systems such as medical robots need 
explicit control of the workflow with clear expert-driven definition of the status path. This can be addressed by Finite State Machine (FSM), a formal representation that can be used to model and describe complex discrete-event systems \cite{harel1987statecharts}. FSMs explicitly define different states of the system and the transition between them \cite{zadeh1973outline, harel1987statecharts, grenander1994representations}. In a process-controlled workflow, the completion of a step is formalized as a status  \footnote{\label{ft:state}In this article, ``status'' refers to a generic status of a system, and ``state'' refers to a state defined in FSM terminology.} of a discrete event with multiple possible outcomes. The start, end, and outcome of a state is used to outline system interactions. In this scheme, FSM helps translate human experience into machine-friendly instructions. Under the representation scheme of FSM, medical robotic applications can include hierarchical processes, where tasks are performed concurrently or subdivided into smaller tasks, requiring the use of Hierarchical FSMs (hFSMs). hFSMs \cite{sklyarov1999hierarchical, alur1999communicating, alur1998model} have been widely used in the research of automation, such as autonomous vehicles \cite{kurt2013hierarchical}, smart retail \cite{trinh2011detecting}, or pedestrian behavioral modeling \cite{kielar2014concurrent}. The construct of hFSM may include other FSMs and it can simplify the design of the complicated FSMs and may be a better fit for medical robotic applications.

Moreover, differences between medical procedures cause limited generalizability of medical robot functions. As a result, previous research and commercial platforms have focused on developing robotic system architectures that are applicable to specific medical procedures, albeit the shared features of perception, visualization, planning, and actuation. Perception - the capability of comprehending the surrounding environment of a robotic system, and the ability to locate target anatomy - is crucial for medical and surgical purposes. In medical applications, robot perception is usually accompanied by visualization modules, which assist in navigating the robot for Image-Guided Interventions (IGI). The system can retrieve geometric/anatomic information using medical imaging devices \cite{masamune1995development, loser2000new, cleary2001state, hempel2003mri, fischer2008mri, monfaredi2018mri, rich2021mri}, while incorporating medical image analysis algorithms to assist robot navigation \cite{vannier1996three, jannin2002validation, gao2021fluoroscopic, farvardin2021biomechanically, unberath2021impact, opfermann2021feasibility}. Medical robotic systems also integrate additional sensors within the actuation components to generate robot status data, or track the robot independently of the anatomy \cite{song2011development, henken2013error, taffoni2013optical, khan2019multi, sefati2020data}. In addition to imaging, sensing, visualization, data processing, and actuation, a fully-integrated robotic system for IGI should have a user interface (UI) to take human input, and a simulation platform to support decisions throughout the planning, execution, and updates of the medical plan during the procedure. As the system complexity grows, integration of the perception, actuation, and visualization becomes non-trivial both at the level of the system architecture and the implementation of the procedure. At the moment, each medical or surgical procedure may require a complete re-development and re-integration cycle, even though only the medical workflow needs to be changed. To tackle such complex integrations in medical robots, existing software, such as Surgical Assistant Workstation (SAW), \cite{kazanzides2014open} adopts a distributed modular architecture. For general robotics, Robot Operating System (ROS) uses a multi-process, and peer-to-peer structure \cite{quigley2009ros}. Both systems have a similar philosophy and design goals of supporting integrative research. They achieve this by facilitating the separation of development concerns and building fault-tolerant integration (without single-point of failure) using sub-modules. However, the modular structure of these designs merely holds perception, actuation, and visualization components together with data transmission. For process control, the workflow should also be considered during the system architecture design. 

Recent robotic research for medical procedures focuses separately on the target planning, tool design, and accuracy of the robotic motion \cite{liu2022inside, lancaster2004evaluation, matthaus2006planning, noccaro2021development, ginhoux2013custom, meincke2016automated, farvardin2021biomechanically, basafa2015subject}. Here we take a holistic view combining the potential engineering improvements of robotic systems and consideration of the integrity of medical workflows. Our proposed paradigm also aims at fast reconfiguration, redevelopment, and re-integration of a robotic system for different applications. The deterministic steps are automatable with more software development efforts. Our contributions can be summarized as the following: 
\begin{itemize}
    \item A process-controlled medical robotics system (PCMRS) that helps with customization for controlled workflow using hFSM.
    \item A central philosophy of a modular system in medical robotics, to provide a structured way to design process-controlled medical robotic systems.
    \item Shown in Section \ref{sec:methods}, the system can lead to a fully automatic paradigm, by identifying the standardized steps that are deterministic in the design cycle.
    \item We use Robot-Assisted Transcranial Magnetic Stimulation \cite{liu2022inside} as an example for a neurology application, and Robot-Assisted Femoroplasty as an example for establishing a new procedure in orthopedic surgery \cite{bakhtiarinejad2023surgical}. 
    \item We performed fault injection experiments to test workflow integrity, where we introduced artificial errors and observed system responses. We also carried out human subject tests and robotic tool placement experiment on phantoms.
\end{itemize}

\section{Methods}\label{sec:methods}

\begin{figure}
\centerline{\includegraphics[width=0.9\columnwidth]{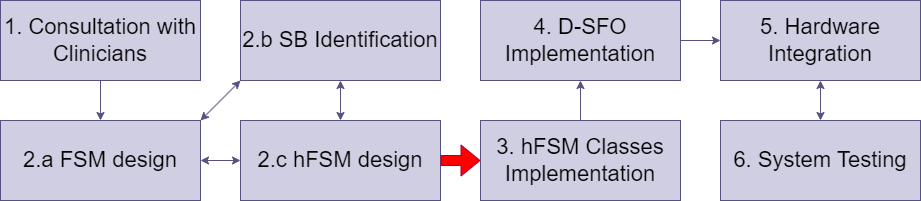}}
\caption{Design cycle of a PCMRS system. FSM design, SB (State Branch) identification, hFSM design, and hFSM classes implementation are introduced in Section \ref{sec:hFSM}. The red arrow indicates the conversion from hierarchical FSM to software implementation, which is deterministic and thus can be automated. The conversion is a process that translates hFSM into codes. D-SFO (DispatcherState/Flag/Operation paradigm) implementation is introduced in Section \ref{sec:d-sfo}. In future, the step 1 can be an automated process as shown in ongoing research in generative models \cite{strong2023chatbot}. Subsequently the design cycle can be fully automated.}
\label{fig:PCMRSprocedure}
\end{figure}

Figure \ref{fig:PCMRSprocedure} shows the design process of PCMRS. The procedure starts with a consultation with clinicians, ensuring an understanding of the requirements and needs for the medical procedure. Based on the consultation, the engineering team designs the FSM, effectively translating the medical task to engineering procedures (Section \ref{sec:hFSM}). To simplify the FSM design, State Branches (SB) (Section \ref{sec:imprules} Definition \ref{def:SB}) that represent a set of tasks with interdependency, are identified and hFSM is generated. The hFSM  can then be automatically converted to codes for software implementation (Section \ref{sec:softwareimplementation}), and integrated into a Dispatcher-State/Flag/Operation paradigm (Section \ref{sec:d-sfo}). The design cycle is then completed with hardware integration and system testing. The process of FSM design, SB identification, and hFSM design can be iterative. If the medical procedure has already been standardized by the medical community, the hFSM design is also deterministic and can be automated. Recent research on specialized generative models has shown such potential \cite{strong2023chatbot}. Thus, fully automation end-to-end in the design cycle is possible.

\begin{figure}[ht]
\centerline{\includegraphics[width=0.85\columnwidth]{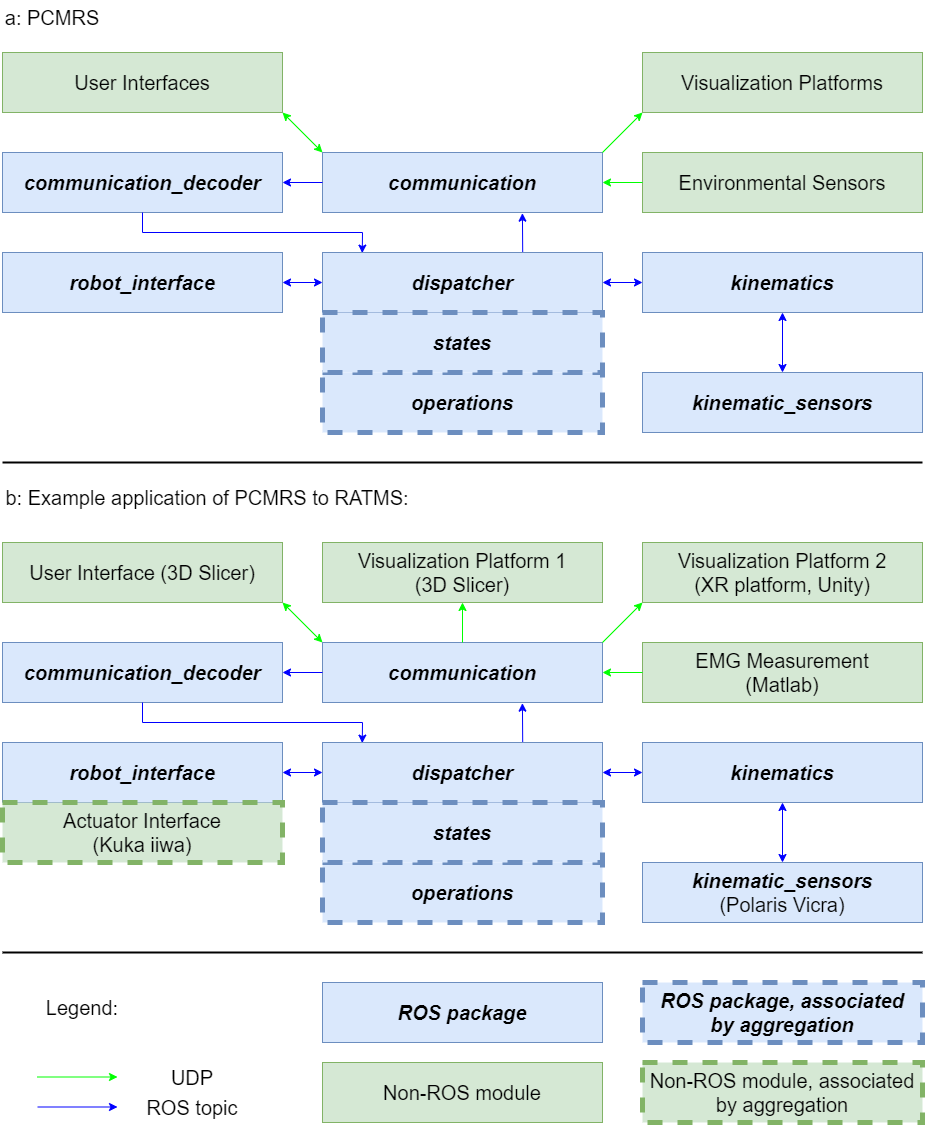}}
\caption{PCMRS system diagram. This figure illustrates the organization and communications between software modules in a PCMRS system with minimal  extensions used for RATMS. Boxes in blue with bold text are modules implemented using ROS packages in C++. Boxes in green are modules implemented using non-ROS software. Specifically, 3D Slicer module is the stand-alone application with customized Python scripting, Unity module is the Unity platform with C\# scripts, and the Kuka iiwa robot interface is developed in C++ libraries. 3D Slicer is an example of planning and visualization module, and can be replaced by any software with TCP/UDP interfaces. Boxes with dashed lines denote classes with weak ownership   that are associated by aggregation (Unified Modeling Language terminology) with those in boxes with solid lines. Green arrows denote connections in UDP, and blue arrows are in ROS topics. Note that each blue box is a ROS package containing several ROS nodes, and each blue arrow may indicate several ROS topics.}
\label{fig:PCMRSsys}
\end{figure}

\begin{figure}[ht]
\centerline{\includegraphics[width=0.85\columnwidth]{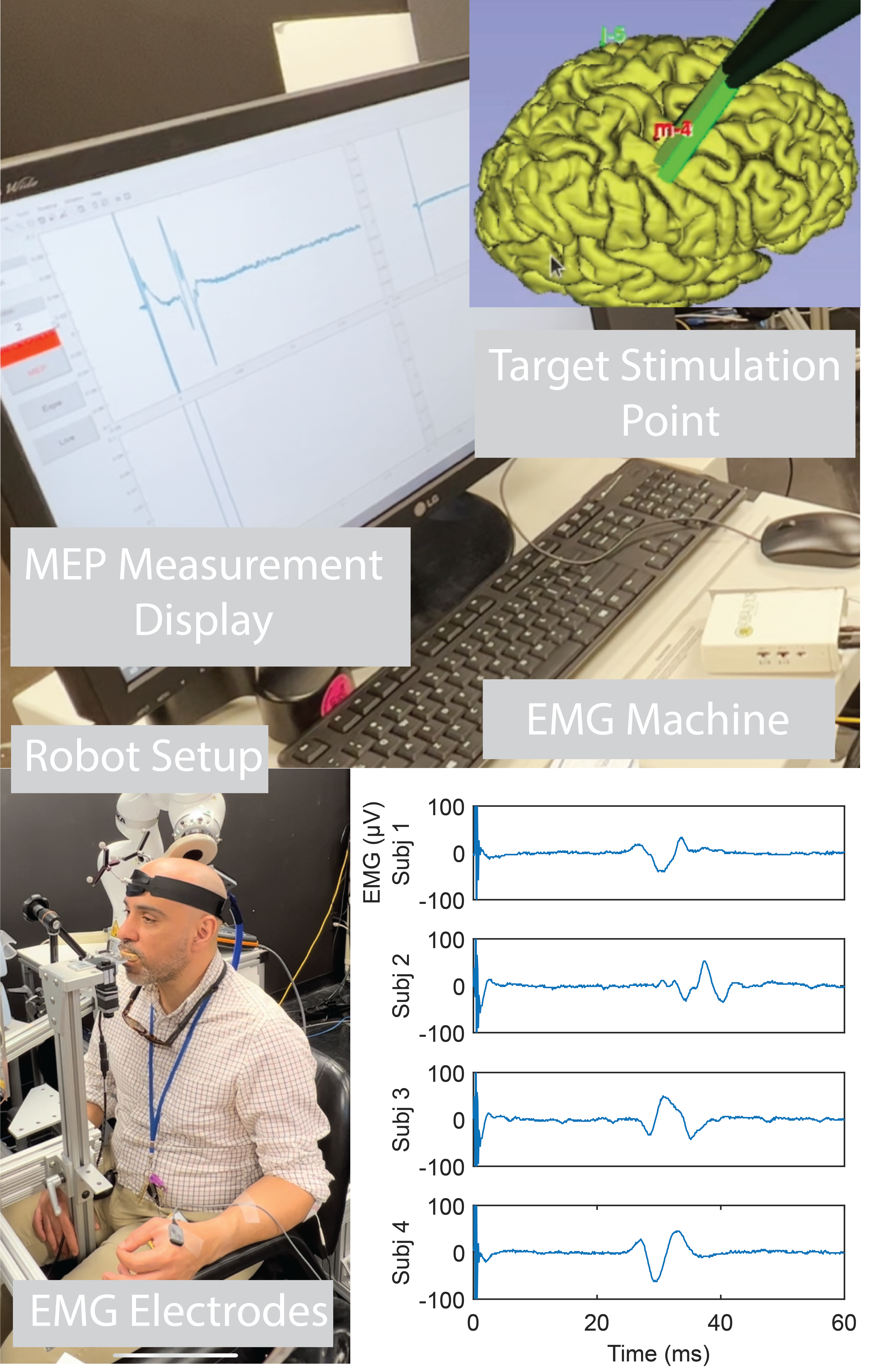}}
\caption{The RATMS setup. In this procedure, Electromyography (EMG) is used to measure motor evoked potential (MEP) in response to TMS pulses delivered at the motor cortex. The target stimulation point is at the hand knob within the primary motor cortex \cite{lemon2008descending}. The example EMG waveforms with TMS are shown from 4 different subjects. The EMG signal is measured with electrodes attached to the first dorsal interosseous (FDI) in the hand contralateral to the side of TMS application (i.e, left hand with TMS at the right motor cortex).}
\label{fig:MEP}
\end{figure}

Fig. \ref{fig:PCMRSsys} shows a system with minimal extensions using PCMRS and its application to Robotic Assisted Transcranial Magnetic Stimulation (RATMS). It is an integrated system with actuation, perception, and visualization modules. Some modules are application-specific; for example, the EMG measurement module (Fig. \ref{fig:MEP}), the Kuka iiwa robot and the Polaris Vicra optical tracker. The integration of the modules is by ROS topics, transmission control protocol (TCP), and user datagram protocol (UDP). The ROS topics are used when the modules are implemented in ROS, and the TCP/UDP protocol is used when the modules are implemented  with extensions or hardware that do not have ROS-ready interfaces in the driving   software. This integration merely holds all the modules in the system   together, similar to existing systems, as discussed in the introduction section. Thus, we add a D-SFO paradigm that enables process control capability in the integrated system as discussed in Section \ref{sec:d-sfo}.

The centerpiece of PCMRS is the \textit{dispatcher} module 
\footnote{In this section, if not explicitly noted, the term ``module'' is used to refer to a subentity of PCMRS on the system level. ``Component'' is a generic reference to describe a constituent part of a larger entity .}, 
which manages the processing of data and commands. The UDP communication modules \textit{comm} and \textit{comm\_decode} manage the non-ROS data I/O, then send the decoded data to the \textit{dispatcher}. The \textit{dispatcher} also contains \textit{states} and \textit{operations} modules, which correspond to the hierarchical FSM and state operations 
\footnote{In this section, if not explicitly noted, an ``operation'' refers to a State Operation (SO), defined in Section \ref{sec:imprules} Definition \ref{def:SO}, and an ``action'' refers to a generic subroutine or procedure, with no specific distinction.}
. The \textit{kinematics} module carries out the kinematics calculations for visualization, data recording, accurate placement of the robot or other spatial transformation applications. ROS-compatible sensors can be connected to the \textit{kinematics} module or the \textit{dispatcher} module, depending on whether the sensor data is coupled with kinematics calculation.

\begin{figure}[ht]
\centerline{\includegraphics[width=0.8\columnwidth]{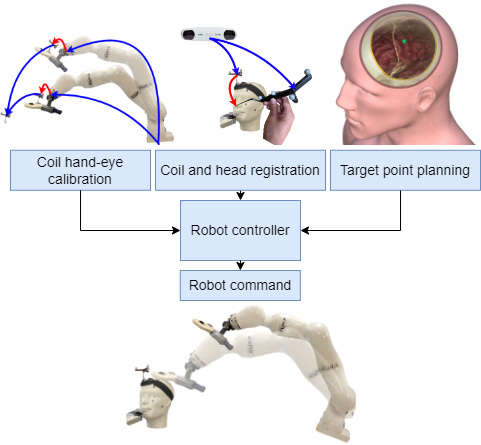}}
\caption{Uncontrolled workflow for an RATMS system \cite{liu2022inside}. The directions of the arrows in the figure represent data flow. }
\label{fig:oldworkflow}
\end{figure}

Fig. \ref{fig:PCMRSsys}b shows an example case where an optical tracker is connected to the \textit{kinematics} module. Our RATMS system relies on an optical tracker \cite{liu2022inside} to retrieve the spatial information of the TMS coil and subject’s head (Section \ref{sec:hFSM} and Fig. \ref{fig:oldworkflow}). Similarly, Robot-Assisted Femoroplasty also depends on optical tracking to obtain the poses of the drilling and injection guide (Fig. \ref{fig:FemurSetup}) and subject's femur \cite{farvardin2021biomechanically, bakhtiarinejad2023surgical}. Other ROS-compatible hardware and software can be integrated by ROS topics, services, or actions.

\subsection{FSM in Medical Procedures}\label{sec:hFSM}

We propose to use hierarchical FSM for medical robotic procedures. This section expounds on the rationale for using a controlled workflow and describes the methods used for the design and implementation of the hierarchical FSM. Here we use RATMS as a demonstrative case and propose possible extensions.

TMS is a neuro modulation technique based on electromagnetic induction of electric current inside the brain \cite{barker1985non}. In this process, RATMS can enhance the accuracy of TMS coil placement and facilitate the procedure \cite{richter2013optimal, lancaster2004evaluation, matthaus2006planning}. The RATMS workflow (without process control) contains the steps required to (1) calibrate the TMS coil and robot end-effector, (2) register the head to the MRI from the subject, (3) plan a target location on the MRI for TMS application and (4) direct the robot to align the TMS coil with the target location on the head. The RATMS setup and workflow are shown in Fig. \ref{fig:MEP} and Fig. \ref{fig:oldworkflow}, and further details for implementation are provided in \cite{liu2022inside}. Previous studies have focused on improving the robotic approach compared to the manual approach with respect to the accuracy of the tool placement, the effect of stimulation, and finding a target location within the brain, but not on the control of the workflow \cite{noccaro2021development, ginhoux2013custom, meincke2016automated, harquel2017automatized, yi2010design, lancaster2004evaluation}. The uncontrolled workflow of RATMS depends heavily on real-time human decisions, whilst a controlled workflow is pre-defined and automated, considering human operators as an integral component of the system. 

\subsubsection{FSM}

\begin{figure}[ht]
\centerline{\includegraphics[width=\columnwidth]{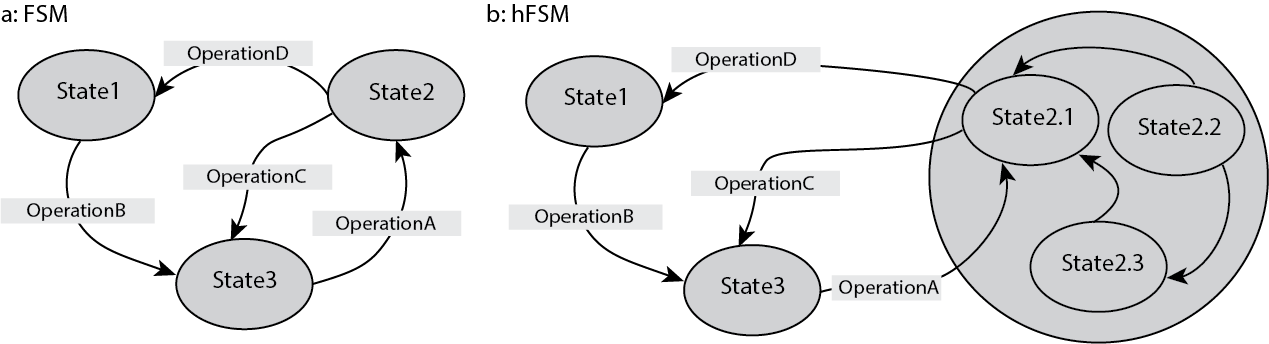}}
\caption{Fig. a. An example FSM containing 3 possible states and 4 possible operations. The system starts at State 1 and transitions to State 3 by completing Operation B. Successful Operation A and Operation C cause alternative state switching between State 2 and State 3, following the arrows connecting the states. At State 2, the successful completion of Operation D causes a transition back to State 1. The system will forbid any other operations not depicted in the diagram, and any failed execution of the depicted operations will not lead to a state transition. Fig. b. A simple hierarchical FSM. The State 2 in a. is expanded to be another FSM.}
\label{fig:simpleFSM}
\end{figure}

FSM is commonly used in process control \cite{sklyarov1999hierarchical, drumea2004finite}. A simple example of FSM is shown in Fig. \ref{fig:simpleFSM}a, where a blueprint of three states and four possible transitions are shown. In Fig. \ref{fig:simpleFSM}a, state transition only happens when an operation defined in the diagram is executed successfully. Any failed operation will not cause a state transition, and any operation not depicted will be forbidden by the system. Thus, the security of the workflow is maintained without any online human decisions. A simplified FSM using a hierarchical structure that allows parallel and intermediate processes is shown in Fig. \ref{fig:simpleFSM}b, where the original State 2 in Fig. \ref{fig:simpleFSM}a is isolated out as an additional FSM and operates separately.

\begin{figure*}[ht]
\centerline{\includegraphics[width=\textwidth]{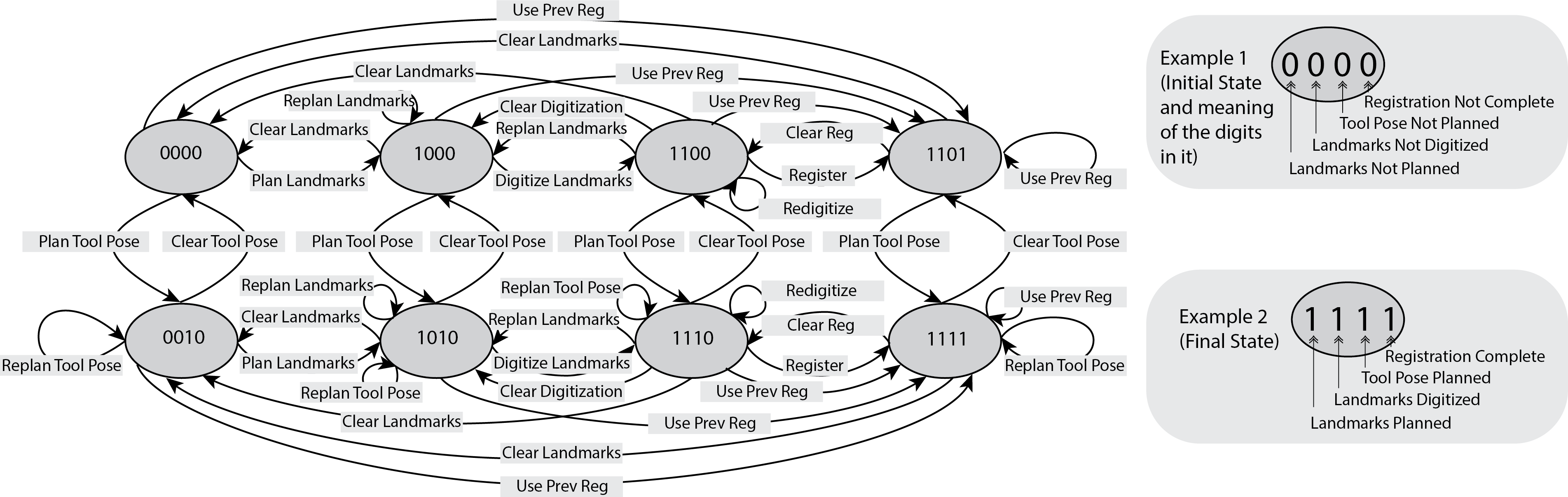}}
\caption{Single level FSM for RATMS (non-hierarchical). The four digits inside a state denote the statuses of four critical operations, shown in the two examples (initial state and the final state) at the right side of the figure. Each critical operation is a binary flag-setting action. The maximum number of possible statuses is $2^N$, where $N$ is the number of critical operations. However, in this figure, only 8 states are possible because of the dependencies of the steps: ``Registration completed'' is impossible when landmarks are not planned or not digitized, and the ``landmark digitized'' status depends on the ``landmark planned'' status. A multi-level FSM design (hierarchical) can be seen in Fig. \ref{fig:FSM-RATMS2}, where the registration process is isolated as a level 1 state branch. Additional state branches are also possible. For example, a landmark digitization branch, a robot connection branch, and a robot control branch. Fig. \ref{fig:FSM-RATMS2} significantly simplifies the seemingly complex states transitions in Fig. \ref{fig:FSM-RATMS1}.}
\label{fig:FSM-RATMS1}
\end{figure*}

We have designed an FSM diagram for RATMS with critical operations including planning landmarks, digitizing landmarks, planning placement poses, and performing registration (Fig. \ref{fig:FSM-RATMS1}). The uncontrolled workflow of RATMS as shown in Fig. \ref{fig:oldworkflow}, has no explicit order of operations, and the transitions of steps are highly dependent on human decision. Therefore, status of the system workflow is not automated and monitored, making it susceptible to human errors. Using flags can meet the minimum requirement of the indications of the system status (discussed in Section \ref{sec:introduction}). However, a explicit blueprint of possible states and operations is needed to automate the processes and maintain security (For example, the robot is forbidden to move, unless prior operations are successfully completed, as discussed in later sections). In our framework, the non-hierarchical FSM can be useful when there is a limited number of critical operations. The system status entirely depends on completing all critical operations, where each critical operation can be seen as a binary flag-setting action. The maximum required number of states in this case is $2^N$, where $N$ is the number of possible critical operations. However, note that while designing states, some may not possibly exist. For example, ``0100'' does not exist because it means ``landmarks are not planned, landmarks are digitized, placement pose is not planned, and registration is not complete''. ``Landmarks are not planned, landmarks are digitized'' are in conflict, and thus not a possible state. 

\begin{figure*}[ht]
\centerline{\includegraphics[width=0.8\textwidth]{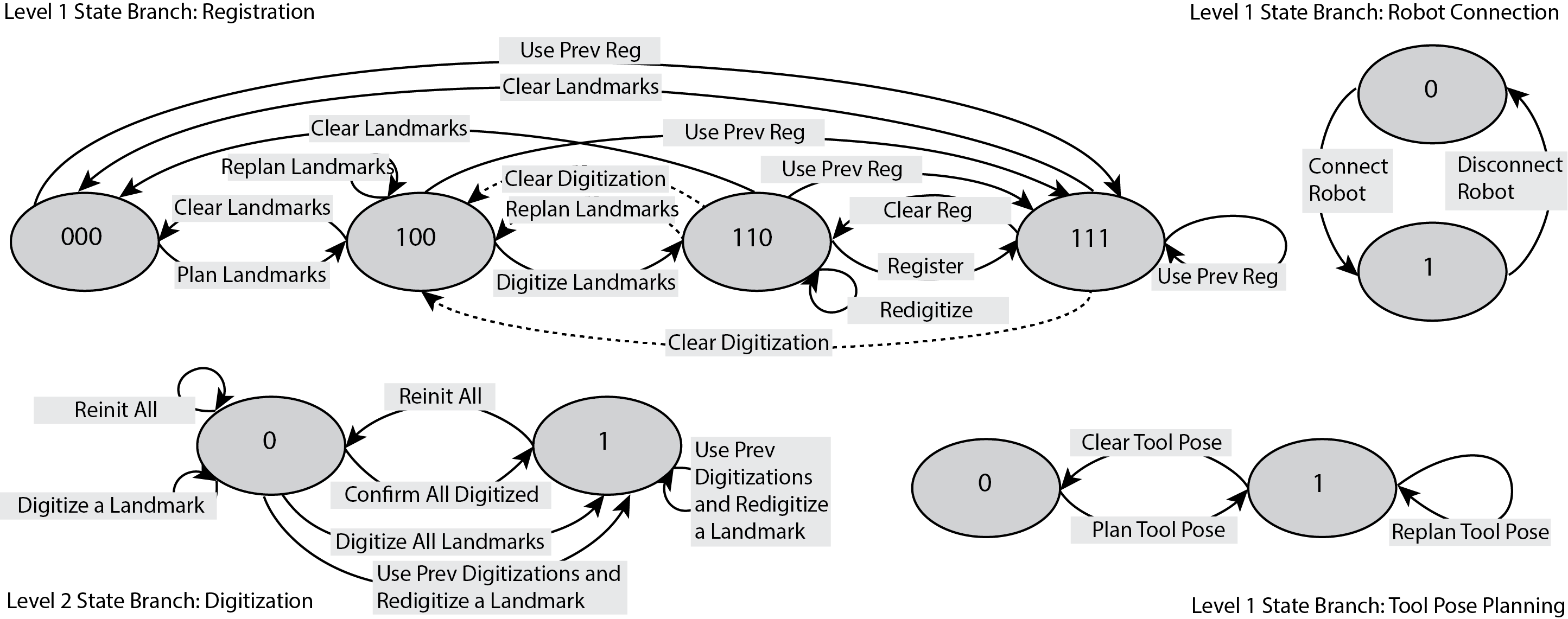}}
\caption{Hierarchical FSM for RATMS. All states at all levels should have a reinitialization operation. Most are not shown in the diagram for simplicity. The registration branch contains a level 2 branch (digitization) at 100 to 110. Any level 2 state change (from one state to another) must emit a signal to the corresponding level 1 state, as defined in Rule \ref{thm:SCO}. Any DIO to 100 (dash lines) should call the reinitialization state of the digitization branch, as defined in Rule \ref{thm:DIO}. }
\label{fig:FSM-RATMS2}
\end{figure*}

To make the system more compact, Fig. \ref{fig:FSM-RATMS2} isolates some states as individual FSMs, showing a hierarchical FSM. The ``registration'' process by itself can be a self-contained workflow, which we call a state branch (as in State Branch, Section \ref{sec:imprules} Definition \ref{def:SB}). ``Digitization'' is also a branch, on which ``registration'' process depends. Isolating the states in this manner,  not only makes the original FSM simpler, but also adds additional capabilities for the ``digitization'' process. For example, in RATMS, some digitizations of the planned landmarks may be suboptimal, so it is ideal to re-digitize these planned landmarks. The isolated ``digitization'' branch makes this possible as a self-contained state branch. Furthermore, additional parallel levels and hierarchies can be added, such as robot connection status (branch C in Fig. \ref{fig:FSM-RATMS2}) and robot control status (not shown in Fig. \ref{fig:FSM-RATMS2}). 

\subsubsection{hFSM Implementation Rules}\label{sec:imprules}

The states within a hierarchical FSM must be updated in a structured manner. Therefore, we establish definitions and rules for configuring an FSM for workflows to ensure a systematic approach.

\begin{thm2}
\label{def:SB}
State Branch (SB): An FSM representing the state changes and operations in a self-contained workflow.
\end{thm2}

\begin{thm2}
State Level: An SB can have state digits representing the states of another SB (hierarchy). A State Level is a different numeric value to represent the levels of branches, or their ``depth'' within the hierarchy. The State Level starts from 1, representing a root workflow.
\end{thm2}

\begin{thm2}
Re-initiation Operation (RIO): An operation that transitions a state to the starting state of an SB.
\end{thm2}

\begin{thm2}
\label{def:SO}
State Operation (SO): An operation that points from one state to another. It can be a State Changing Operation (SCO) (an operation that points to a state from another different state), or a State Maintaining Operation (SMO) (an operation that points to a state from the same state itself).
\end{thm2}

\begin{thm2}
Incoming Operation: From the perspective of a state, Incoming Operation is pointing toward the state. It can be a Direct Incoming Operation (DIO) (an operation pointing toward a state from another state at the same N State Branch, and the operation \textbf{is not called} by an N+1 SB) or an Indirect Incoming Operation (IIO) (an operation pointing toward a state from another state at the same N State Branch, and the operation \textbf{is called} by an N+1 SB).
\end{thm2}

\begin{thm}
Any state at a State Level of $N > 1$ should have a re-initiation Operation.
\end{thm}

\begin{thm}\label{thm:SCO}
Any SCO at a State Level of N+1 should emit a signal to update the level N branch by calling an operation at the level N branch (IIO).
\end{thm}

\begin{thm}\label{thm:DIO}
Any DIO, pointing to a state that is at level N and is the starting state of the corresponding N+1 branch, should call the re-initiation operation of that N+1 level branch. This rule does not apply to IIO to avoid infinite callback loops, because otherwise by Rule \ref{thm:SCO}, the reinitialization at N+1 will emit a signal which triggers the same IIO on N.
\end{thm}

\subsubsection{hFSM Software Implementation}\label{sec:softwareimplementation} 

\begin{figure}[ht]
\centerline{\includegraphics[width=0.95\columnwidth]{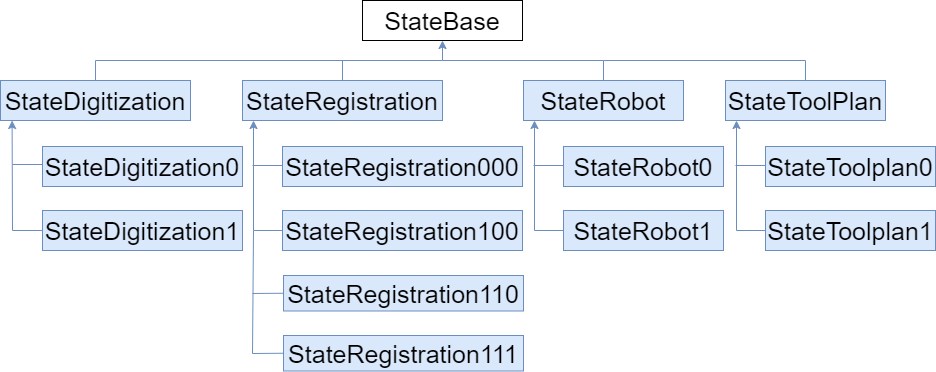}}
\caption{Class inheritance of FSMs. Four SBs are shown here: digitization, registration, robot connection, and tool pose planning. Each SB contains the possible states as the derived classes of their corresponding SB base class. The class name is followed by the state digits indicating the activation (completion) status of the corresponding action.}
\label{fig:FSMclasses}
\end{figure}

The hierarchical FSM can be implemented using class inheritance feature in Object-Oriented Programming (OOP), allowing derived classes to inherit all methods and attributes from the parent class. The inheritance diagram of the hierarchical FSM in PCMRS-RATMS is shown in Fig. \ref{fig:FSMclasses}. Inheritance facilitates the design of FSM by the feature of function overriding.   Specifically, a parent class can represent an SB, and a derived class can represent a state within the SB. An operation in an SB is written as a member function in the class. The parent class of an SB contains all possible operations in all the states of the branch, but a derived class only contains the valid actions at the state it represents. The member functions of the derived class override the corresponding ones in the parent class. All member functions of the parent class are implemented as invalid operations: If the member function is not overridden in a derived class, the method in the parent class will be called, representing an invalid operation. Thus, it is rejected by the system from proceeding.

\begin{figure}[ht]
\centerline{\includegraphics[width=0.8\columnwidth]{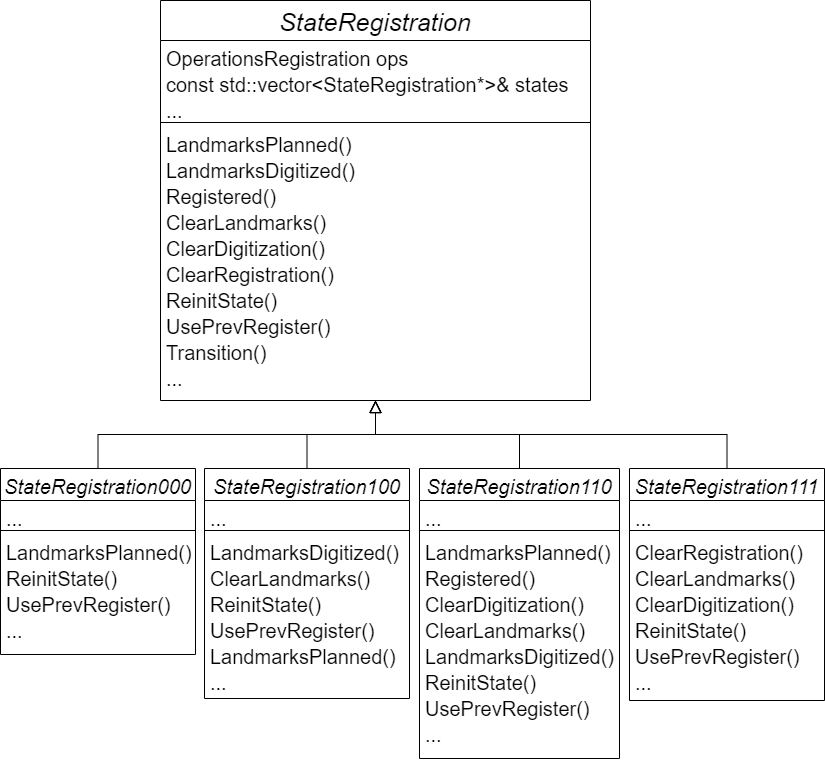}}
\caption{Unified Modeling Language (UML) diagram of the ``registration'' SB. Not all member attributes and member functions are shown. The overriding of the functions ensures execution of allowed state transitions or rejection of forbidden operations. Take \textit{Registered} as an example of a forbidden operation, \textit{Registered} function is not overridden in the class \textit{StateRegistration000}, because it is not possible to register at this state. When calling \textit{Registered} function at state ``0000'', the default method \textit{Registered} in \textit{StateRegistration} (base class of the ``registration'' SB) is invoked, which represents an invalid operation, so the state transition is not executed. The class designs of other SBs use the same principle and are not shown in this paper. }
\label{fig:regFSM}
\end{figure}

As an example, Fig. \ref{fig:regFSM} illustrates the class design of the FSM for the registration process (Section \ref{sec:registration_state_machine}). The ``register'' action can only be performed after ``landmarks are planned and digitized'' (first and second digits in the three-digit notation of the FSM). The states ``100'' and ``110''  in this SB are represented by the class \textit{StateRegistration100} and \textit{StateRegistration110}, respectively. ``100'' denotes ``the landmarks are planned at the state'', and ``110'' means ``the landmarks are both planned and digitized at the state''. ``Register'' is a valid action at state ``110'' but is invalid at state ``100''. This means when the \textit{Register} member function is called by the class \textit{StateRegistration100}, the default member function in the parent class \textit{StateRegistration} is executed, representing invalid action. The registration action is handled as unsuccessful. On the contrary, when the ``Registered'' member function is called by the class \textit{StateRegistration110}, the registration action can be successfully processed. Additionally, using static class or singleton design patterns to implement states can ensure that any state is only instantiated once, improving state security. 

\subsubsection{Registration - A Level N FSM} \label{sec:registration_state_machine}

The registration process in RATMS consists of identifying, digitizing, and registering landmarks. To obtain a good registration, the process may need to be repeated. The repetition mainly involves re-identifying and re-digitizing some of the landmarks, which requires resetting the states of the landmarks. An FSM clearly defines the dependency of the steps. It ensures that an invalid operation cannot proceed if a requirement is not satisfied, as described in Section \ref{sec:softwareimplementation}. The implementation details are shown in Fig. \ref{fig:regFSM}.

\subsubsection{Pose Planning - A Parallel Level N FSM}

The pose plan SB is separated from the original single-level FSM (Fig. \ref{fig:FSM-RATMS1}). Planning the pose can be seen as a parallel branch with registration because the registration process does not depend on the pose planning. In this case, pose planning becomes a binary-state FSM, only having ``complete'' and ``incomplete'' states, with the option of ``re-plan'' that loops the ``complete'' state.

\subsubsection{Digitization - A Level N+1 FSM}

Since there are multiple landmarks in a registration procedure, re-digitizing any of the landmarks invokes a transition from ``all landmarks digitized'' to ``digitization not complete''. Therefore, the digitization SB is extended from the original single-level FSM (Fig. \ref{fig:FSM-RATMS1}). The single-level FSM still maintains a state of the digitization process but is only set to true when all landmarks are digitized. Effectively, the state ``1'' of the digitization SB is the same as the state ``2.1'' in Fig.  \ref{fig:simpleFSM}, which is the handle of the upper-level FSM.

\subsubsection{Robot Connection - A Parallel Level N FSM}

Robot FSM is used to monitor the state of the robot connection. It is also on the same SB level as the pose planning SB and registration SB, so their SOs can be processed in parallel. 

\subsubsection{Other Possible FSMs} In a more complex case, a procedure may depend on a series of consecutive robot positions, forming a defined trajectory. For example, a TMS experiment may involve the mapping of the functional areas \cite{ammann2020framework, harquel2017automatized, meincke2016automated}, where a grid of stimulation points are defined on the cortex, and responses measured after each stimulation. In such cases, the FSM branch can be designed so that the stimulation points are executed in a pre-defined order, and failed operation in a preceding stimulation point prevents the robot from progressing to the next stimulation point.

\subsection{Dispatcher-State/Flag/Operation Paradigm}\label{sec:d-sfo}

\begin{figure}[ht]
\centerline{\includegraphics[width=\columnwidth]{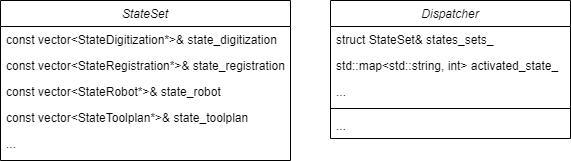}}
\caption{UML diagram  of the D-SFO. Not all member attributes and member functions are shown. \textit{StateSet} is a class to keep a record of activated states from all SBs. Each SB class contains its corresponding set of \textit{operations} (Fig. \ref{fig:regFSM}) and \textit{flags} (not shown).}
\label{fig:dispatcher}
\end{figure}

PCMRS adopts a dispatcher-state/flag/operation (D-SFO) paradigm. The \textit{dispatcher} associates the \textit{states} by aggregation and is a processing module to transition \textit{states}, check \textit{flags}, and call valid \textit{operations} invoked by the user. The class designs of the ``registration'' SB are shown in Fig. \ref{fig:regFSM}, and the class design of the \textit{dispatcher} is shown in Fig. \ref{fig:dispatcher}.

The \textit{operations} module is the collection of subroutines that are command processing instructions and abstractions of physical operations. State Changing Operations (SCOs) and State Maintaining Operations (SMOs) have been encapsulated in the \textit{operations} module. The D-SFO paradigm isolates the execution of subroutines from the processing of received commands and the maintenance of states, ensuring the FSM transitions happen entirely based on the execution result of SCOs and SMOs.  A state-relevant command is an operation that can invoke an SCO or an SMO. The dispatcher regards the command as a request for a validity check for state transition. A state-relevant command can be directly sent by the user or by an intermediate operation. Sending the request directly by the user can be by clicking a button within the user interface, and sending by an intermediate operation can be a timer executing a state operation (SO). In either case, the dispatcher receives commands and checks the validity of the operation at the current state, then decides whether to invoke an SO. If the operation is valid at the current state, then the SO is executed first, followed by a state transition. Otherwise, the dispatcher rejects the operation request and will not execute the SO. In the implementation, an operation entity is a C++ function that executes a sequence of subroutines predefined in the \textit{operations} module. The execution of the subroutines is done by calling the member function  \textit{Transition} in a state class (Fig. \ref{fig:regFSM}). Algorithm \ref{alg:operation} shows how a state class defines the execution.

The implementation of the \textit{dispatcher} uses ROS topics, adopting the observer-callback design pattern. When \textit{dispatcher} receives a message that evokes a callback, the corresponding SB executes an operation and a state transition. Algorithm \ref{alg:dispatchercallback} defines a callback.

\begin{algorithm}[ht]
\caption{Execution of an operation in a state. $f_i$ is a step required to complete an operation. $states$ is the member attribute in the \textit{State} class that holds the reference to the states array, which contains all possible states of the SB. $T$ is the target state (The end of an arrow in FSM diagram).}\label{alg:operation}
\begin{algorithmic}[1]
\Procedure{State::Operation()}{}
\State $\textit{S} \gets \textit{\{$f_1,f_2,f_3,...,f_n$\}}$
\State $\textit{t} \gets \textit{target status number}$
\State State::Transition($t,S$)
\EndProcedure
\end{algorithmic}
\begin{algorithmic}[1]
\Procedure{State::Transition($t, S$)}{}
\State State::Deactivate()
\ForAll{$f\in\mathit{S}$}
\State execute $f$
\EndFor
\State $\textit{T} \gets \textit{states(t)}$
\State State::Activate($T$)
\State State::CheckUniqueStateActivated()
\EndProcedure
\end{algorithmic}
\end{algorithm}

\begin{algorithm}[ht]
\caption{Dispatcher callbacks. $D$ is the data passed to the \textit{dispatcher}. $t$ is the target state index.}\label{alg:dispatchercallback}
\begin{algorithmic}[1]
\Procedure{Dispatcher::Callback($D$)}{}
\State  $\textit{d} \gets \text{ Preprocess $D$}$
\State Dispatcher::Process($d$)
\State $\textit{a} \gets \text{ The number of the activated state of the SB}$
\State $\textit{SB} \gets $ Dispatcher.state\_set(\textit{a})
\State $\textit{t} \gets $ SB::Operation()
\State Dispatcher::StateTransitionCheck(\textit{t})
\EndProcedure
\end{algorithmic}
\end{algorithm}

In Section \ref{sec:introduction}, we have argued against using flags for process control. However, the D-SFO paradigm still contains \textit{flags}. The \textit{flags} here are used to keep records of the status of system components by FSM, but not for process control. The \textit{state} transitions will update the \textit{flags} automatically, but the \textit{flags} do not interfere with the state transition. For example, in the SB for registration, one state has a three-digit representation, where each digit shows the status of landmark planning, landmark digitization, and registration completion. The D-SFO here contains three flags only to show the progress of these steps. The execution of the steps depends on the \textit{states}, but not the \textit{flags}. Eliminating a digit in a state in FSM requires rearranging the process control scheme and FSM design, but eliminating a flag does not have consequences unless there is an explicit need for showing the status of the corresponding step. 

\subsection{Communication Module}

The communication modules in PCMRS use ROS topics and UDP protocol. ROS topic is a message transmission method between ROS nodes. ROS also offers other methods for information exchange (services and actions) that can serve different purposes. However, ROS topics are particularly advantageous as they require fewer development efforts and meet most needs for message exchange. The ROS packages shown in Fig. \ref{fig:PCMRSsys} mainly use ROS topics for communication, shown in blue arrows. In addition, PCMRS uses \textit{comm} (stands for communication) to transmit UDP packets with modules that are not implemented using ROS, and it uses \textit{comm\_decode} to decode the incoming UDP packets. The \textit{dispatcher} in the D-SFO paradigm then processes the decoded data. 

\subsubsection{UDP Communication with ROS}\label{sec:rosudp}

ROS provides a wide range of pre-developed packages for robotics, offering a robust and extensively developed messaging system. However, some hardware and software may not have a ROS-compatible interface. We developed \textit{simple-ros-comm}, a string-based commands processing paradigm, which requires the user to customize a pair of command decoders and encoder. It is flexible because the user has complete control over the format and content of the UDP packets. The data used in PCMRS are all 3D coordinates, 6D poses, and 7D joint angles, which are directly appended after a 16 character strings command and decoded in the \textit{comm} module as ROS topics. Algorithm \ref{alg:roscomm} shows how an incoming UDP packet is processed by \textit{comm} and \textit{comm\_decode}, and how the data contained in the packet is distributed to the \textit{dispatcher}.

\begin{algorithm}[ht]
\caption{ROS \textit{comm} and \textit{comm\_decoder} callbacks. $p_{in}$ is the processed data to be sent to the dispatcher for further actions. $i$ is the index of the publisher in the set that holds all publishers intending for different destinations in the dispatcher. The publisher distributes the processed data to the dispatcher.}\label{alg:roscomm}
\begin{algorithmic}[1]
\Procedure{Comm::Callback()}{}
\State  $\textit{p} \gets \text{ Incoming UDP packet}$
\State ${p_{msg}} \gets$ Comm::Process($p$)
\State publisher.publish(${p_{msg}}$)
\EndProcedure
\Procedure{CommDecoder::Callback(${p_{msg}}$)}{}
\State $\textit{$p_{in},i$} \gets \text{ CommDecoder::Process($p_{msg}$)}$
\State publisher $\gets $ CommDecoder.publishers($i$)
\State publisher.publish(${p_{in}}$)
\EndProcedure
\end{algorithmic}
\end{algorithm}

\subsubsection{UDP Communication with 3D Slicer}\label{sec:slicerudp} 

We have developed \textit{simple-slicer-comm}, a UDP communication module for integration with 3D Slicer, which uses the same string-based commands processing paradigm. Non-blocking transmission of data requires parallel processes, but 3D Slicer is a single-threaded application, and the scripted modules do not support multi-threading. So instead, a Qt Timer is used to achieve pseudo-parallel processing without freezing regular functions in 3D Slicer. Algorithm \ref{alg:slicercomm} shows the non-blocking receiving callback function that enables pseudo-parallel processing of single-threaded 3D Slicer.

\begin{algorithm}[ht]
\caption{\textit{simple-slicer-comm} non-blocking receiving callbacks. The callback is called repeatedly by recursion. }\label{alg:slicercomm}
\begin{algorithmic}[1]
\Procedure{SlicerComm::Callback()}{}
\State  $\textit{t} \gets \text{ processing time gap}$
\State  $\textit{p} \gets \text{ Incoming UDP packet}$
\State ${p_{msg}} \gets$ Comm::Process($p$)
\State ScriptedLoadableModule::Logic::Process(${p_{msg}}$)
\State qt::QTimer.singleShot($t$,SlicerComm::Callback)
\EndProcedure
\end{algorithmic}
\end{algorithm}

\begin{figure*}
\centerline{\includegraphics[width=0.8\textwidth]{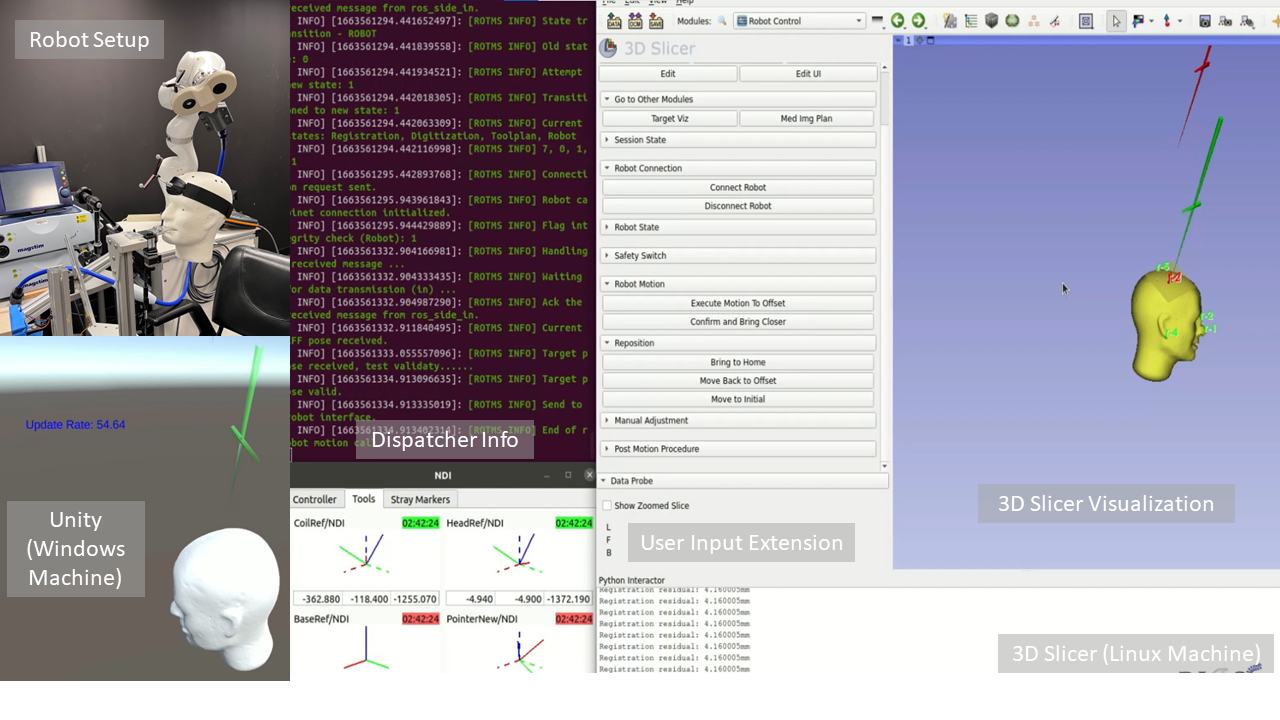}}
\caption{User interface of PCMRS. The top left panel is the setup for  RATMS using a  head phantom. A laser scanner is used to obtain the 3D model of the phantom head for planning and visualization. The top middle panel is the ROS command line, showing the status of the ROS modules. The right panel is the user interface of 3D slicer and the visualization window, hosted on a Linux machine. In the visualization window, the green arrow indicates the planned pose of the TMS coil, and the red arrow indicates the current pose of TMS coil. The bottom left panel is the Unity Editor window, showing the same visualization.  The Unity Editor is hosted by a separate Windows machine, connected by the \textit{comm} module. The frame rate is also shown in the bottom left panel. The bottom middle panel is a CISST-SAW \cite{kazanzides2014open} visualization interface of the Polaris Vicra optical tracker, showing the poses of tracked retroreflective markers (here markers on the head phantom and the TMS coil).}
\label{fig:UI}
\end{figure*}

\subsection{Planning and Visualization Module}

PCMRS achieves visualization in real-time by bridging the 3D Slicer module to the D-SFO system. Unity is used to achieve mixed reality capability in PCMRS. Both 3D Slicer and Unity are non-ROS modules, so the bridging from them to the \textit{dispatcher} module is done by UDP (Section \ref{sec:rosudp} and \ref{sec:slicerudp}).

For RATMS application, the medical image planning module imports MRI and reconstructs 3D head and brain models. First, the planning of the target tool pose is performed on the skin and the brain models. Then data from the planned pose and commands are sent to \textit{dispatcher} through UDP, where it is processed using the methods discussed in Section \ref{sec:d-sfo}, and the processes of the \textit{dispatcher} are controlled by methods discussed in Section \ref{sec:hFSM}. Upon the reception of a visualization command, \textit{dispatcher} triggers calculations  of the kinematic chains in the \textit{kinematics} module (Fig. \ref{fig:PCMRSsys}), and the relative poses of the current tool are sent in real-time to the visualization module of 3D Slicer (or the Unity module, if using the mixed reality capabilities). 

\subsection{Robot, Sensor, and Additional Hardware Modules}

Based on Kuka Sunrise Toolbox (KST) \cite{Safeea2019}, we developed a C++ package, \textit{kuka-sunrise-toolbox-cpp}, for controlling Kuka iiwa. \textit{kuka-sunrise-toolbox-cpp} comprises a Java Sunrise cabinet application and a C++ interface that respectively run on the robot controller and the ROS machine. The Sunrise cabinet application is directly modified from the Java code in KST \cite{Safeea2019}, and the C++ interface is translated from the Matlab implementation of the KST. To connect the Sunrise cabinet and the C++ interface, \textit{kuka-sunrise-toolbox-cpp} uses TCP communication. In PCMRS, the \textit{robot\_interface} module uses \textit{kuka-sunrise-toolbox-cpp}. Unlike other non-ROS modules of PCMRS, \textit{robot\_interface} does not go through \textit{comm} and \textit{comm\_decode}, but is directly bridged to the \textit{dispatcher} module.

For RATMS application, we also use a Polaris Vicra IR camera to track 6D poses of the subject’s head and TMS coil. Additional sensors, actuators, UIs, and visualization platforms can be added by UDP or TCP connections, using \textit{comm} and \textit{comm\_decode} modules. 

\section{Example Experiments and Discussions}

In this section, we discuss the design cycle of PCMRS, and use robotic TMS (RATMS) and robotic femoroplasty (osteoporotic hip augmentation) as examples for testing the system. This section also presents the results of the alignment tasks, failure injection experiments, and the frame rates of the visualization and communication modules. 

\subsection{FSM Design Cycle}\label{sec:designcycle}

Section \ref{sec:hFSM} introduces the design process of a PCMRS. The procedure consists of 1. consultation with clinicians, 2.a FSM design, 2.b SB identification, 2.c hFSM design, 3. hFSM classes implementation, 4. D-SFO implementation, 5. hardware integration, and 6. system validation, shown in Fig. \ref{fig:PCMRSprocedure}. The workflow of the system is not limited to the control of the medical procedures using clinician experience. Surgical scene detection \cite{killeen2023pelphix} and environment awareness can also be integrated in the controlled process. In future, steps 2, 3, and 4 can be automated since the FSM classes (software) are deterministic once hFSMs (blueprint from medical and engineering experts) are designed. However, the automation requires additional software design to ensure an accurate conversion between a graphical representation to C++ codes. Step 1, where the expert knowledge is produced, may also be automated using generative methods. Such potential has been shown in ongoing research \cite{strong2023chatbot}. 

\subsection{Robotic TMS}\label{sec:TMSexample}

\begin{figure}[ht]
\centerline{\includegraphics[width=\columnwidth]{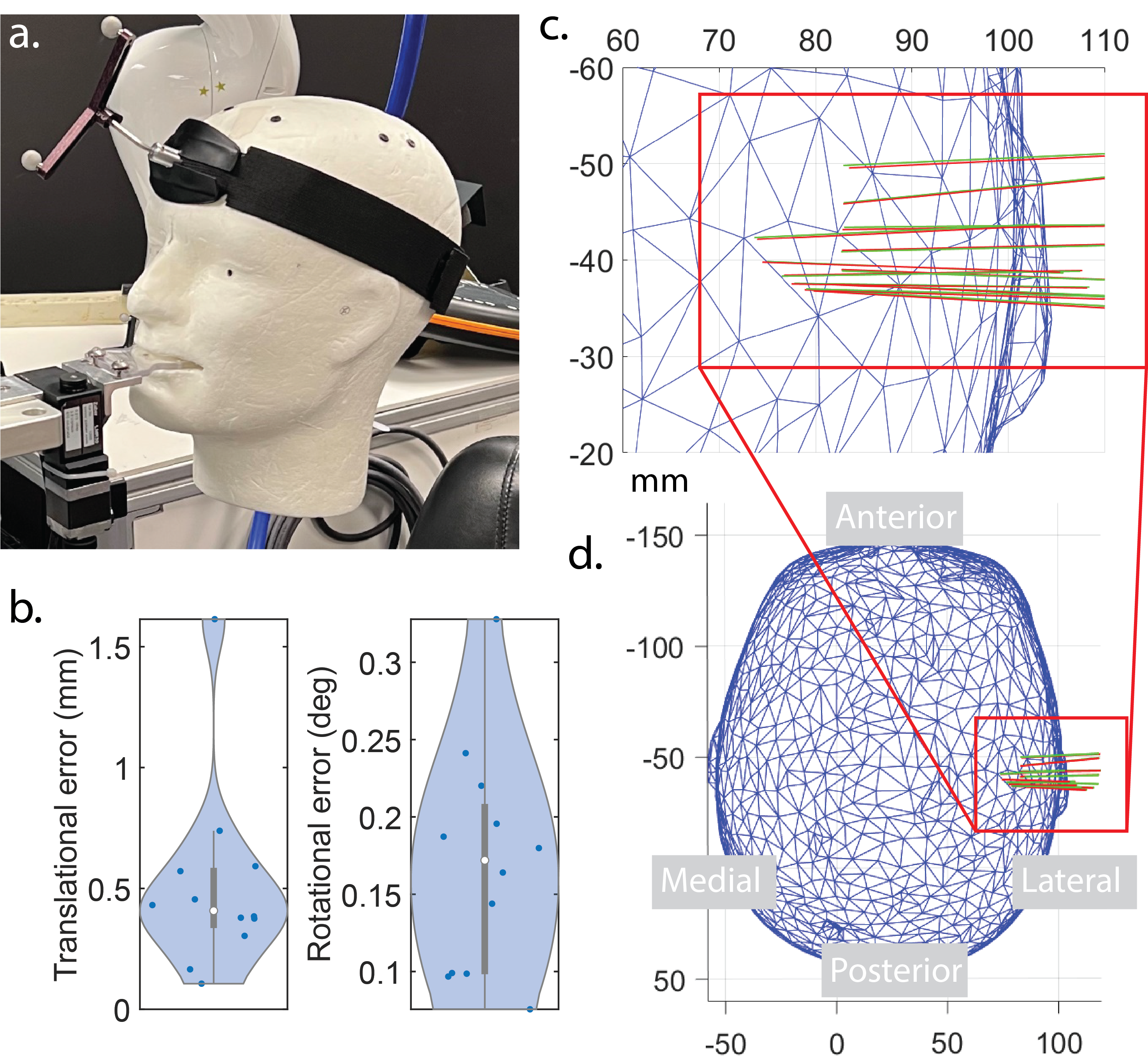}}
\caption{Placement errors in 12 trials for a head phantom used to demonstrate the alignment task in RATMS. Panel a. is the head phantom used in the experiment, which has a similar size of a regular human head. Panel b. shows the errors between the planned tool poses and the measured tool poses. The translational errors are in Euclidean distances, and the rotational errors are angular errors in angle-axis representation. Panels c. and d. contain the blue mesh model of the laser-scanned head phantom (down-sampled tesselation). The green and red straight lines are the 3D representation (vectors) of the planned tool poses and the measured tool poses. The green lines indicate planned tool poses, and the red lines indicate the measured poses after robotic placement.}
\label{fig:placementerrors}
\end{figure}

To test the integration of the software and hardware of PCMRS in RATMS, we tested 12 TMS coil placements on a laser-scanned head phantom.  The experiment procedure was (1) planning 12 different coil poses in the medical image planning module, (2) registering the head phantom, and (3) placing the TMS coil  to the planned pose (steps 1 and 2 can be reversed). The integrated hFSM controls the procedure, and no dependency requirements are violated in the steps or substeps during the experiments. For example, in step 2, landmarks need to be confirmed before digitization, especially after re-planning the landmarks. More details of the workflow design are provided in Fig. \ref{fig:FSM-RATMS2} and Section \ref{sec:hFSM}. The planned and measured poses after all robotic placement are shown in Fig. \ref{fig:placementerrors}. The average translational (Euclidean) error is 0.5096 mm, and the average rotational (axis-angle representation) error is 0.1692 degrees. 

Motor threshold is the minimum strength of the stimulus to evoke a measurable motor evoked potential (MEP) in a muscle. It is an essential measure in experimental and clinical TMS due to the individual differences in the susceptibility to the electrical stimulus. In clinical practice, motor threshold is used as the baseline of the intensity of the applied TMS strength, for both safety and physiological reasons \cite{schutter2006standardized}. Most commonly, the response of single pulse stimulus applied to the primary motor cortex, evoking a motion in the contralateral hand muscle is used to estimate the motor threshold \cite{wassermann1998risk, kim2021optimal}. MEP is the electrical signals recorded using EMG to detect the electrical activity of muscles. The cortical area, ``handknob'', is used in our experiments to elicit movements in the first dorsal interosseous (FDI) in the contralateral hand. It is commonly accepted that the handknob is the hand motor hot spot (hMHS) \cite{ahdab2016hand}. Hot-spot hunting is used to locate the optimal position and orientation of the placement of the TMS coil. Four subjects have been tested. The subjects are all males (average age 29.3) and were screened to exclude history neurologic, or psychiatric illness. The experiments were approved by the Johns Hopkins institutional review board, and informed written consent was obtained from all subjects.  We successfully delivered TMS stimulation, and were able to measure MEPs from all subjects. 

We conducted failure injection experiments to confirm that the system can reject catastrophic errors propagating in the workflow. We injected artificial errors in 10 experiments regarding robotic TMS workflow. The injected error and system rejection results are shown in experiments \# 7,9-11,15-17 in Table \ref{tab:failinj}. The control group is shown in experiments \# 1-3 (without injected error). In the experiments, we introduced 3 types of errors, including missing landmark plan, missing a landmark, and large digitization error. ``Missing landmark plan'' means the operator does not complete landmark planning, corresponding to the operation $000 \rightarrow 100$  in the ``Registration'' SB in Fig. \ref{fig:FSM-RATMS2}. ``Missing a landmark'' means the operator does not digitize a landmark out of all planned landmarks, corresponding to the operation $100 \rightarrow 110$ in the ``Registration'' SB and the operation $0 \rightarrow 1$ in the ``Digitization'' SB. The missing landmark and the total number of landmarks are denoted in Table \ref{tab:failinj} as \#NUM/TOTAL in the Injected Error column. The ``Large Digitization Error'' means the operator digitizes a landmark with a large error, also corresponding to the operation $100 \rightarrow 110$, but is rejected at state $111$. The results of the failure injection experiments show that the system successfully completes the workflow (control groups \#1-3) and it can reject artificial errors. Therefore, the controlled workflow can define failure criteria such that each operation will only lead to predefined behavior.

\begin{table*}
\centering
\caption{Failure injection experiments showing system rejection of catastrophic errors.}
\label{tab:failinj}
\newcolumntype{Y}{>{\centering\arraybackslash}X}
\begin{tabularx}{0.95\textwidth}{
>{\centering}p{0.4cm}>{\centering}p{0.8cm}>{\centering}p{3cm}>{\centering}p{2.9cm}>{\centering}p{2cm}>{\centering}p{1.7cm}>{\centering\arraybackslash}p{4cm}}
\toprule
Test \# & Workflow & Injected Error Type & Injected Error & Ave. Registration Residual & Corresponding to Operation & Step of Rejection (at State) \\
\toprule
1&   &&&2.3009 mm&N/A&N/A\\
2&TMS&None&None&2.5526 mm&N/A&N/A\\
3&   &&&2.3637 mm&N/A&N/A\\
\midrule
4&   &&&1.2677 mm&N/A&N/A\\
5&Fem&None&None&1.0523 mm&N/A&N/A\\
6&   &&&1.1231 mm&N/A&N/A\\
\midrule
7&TMS&Missing Landmark Plan&Missing Landmark Plan&N/A&$000\rightarrow 100$&Planned Landmarks  (100)\\
\midrule
8&Fem&Missing Landmark Plan&Missing Landmark Plan&N/A&$000\rightarrow 100$&Planned Landmarks  (100)\\
\midrule
9&&\multirow{3}{*}{Missing a Landmark}&Missing Landmark \#1/6&N/A&$100\rightarrow 110$&Digitized Landmarks (110)\\
10&TMS&&Missing Landmark \#2/6&N/A&$100\rightarrow 110$&Digitized Landmarks (110)\\
11&&&Missing Landmark \#3/6&N/A&$100\rightarrow 110$&Digitized Landmarks (110)\\
\midrule
12&&\multirow{3}{*}{Missing a Landmark}&Missing Landmark \#1/4&N/A&$100\rightarrow 110$&Digitized Landmarks (110)\\
13&Fem&&Missing Landmark \#2/4&N/A&$100\rightarrow 110$&Digitized Landmarks (110)\\
14&&&Missing Landmark \#3/4&N/A&$100\rightarrow 110$&Digitized Landmarks (110)\\
\midrule
15&&\multirow{3}{*}{Large Digitization Error}&Landmark \#1/6 x 25mm&7.5900 mm&$100\rightarrow 110$&Registered Landmark (111)\\
16&TMS&&Landmark \#2/6 y 20mm&6.2370 mm&$100\rightarrow 110$&Registered Landmark (111)\\
17&&&Landmark \#3/6 z -23mm&6.0678
 mm&$100\rightarrow 110$&Registered Landmark (111)\\
\midrule
18&&\multirow{3}{*}{Large Digitization Error}&Landmark \#1/4 x 25mm&8.6603 mm&$100\rightarrow 110$&Registered Landmark (111)\\
19&Fem&&Landmark \#2/4 y 20mm&8.2317 mm&$100\rightarrow 110$&Registered Landmark (111)\\
20&&&Landmark \#3/4 z -23mm&6.8120 mm&$100\rightarrow 110$&Registered Landmark (111)\\
\bottomrule
\end{tabularx}
\end{table*}

\subsection{Osteoporotic Hip Augmentation}

Osteoporotic hip augmentation, also referred to as femoroplasty, is a therapeutic  procedure for potential treatment to reduce hip fracture risks, which involves injecting bone cement to the femur of the patient. Robotic system with computer-assisted planning and execution can improve the surgical results significantly. In the procedure, the trajectory starts from the entry point on the greater trochanter surface of the femur and ends at the point of the injection inside the femoral neck \cite{basafa2015subject, farvardin2021biomechanically}. The bone drilling and cement injection in the surgery are also alignment tasks, involving aligning the hand-drill or the injection device to the desired planned location and orientation \cite{bakhtiarinejad2023surgical}. 

\begin{figure}
    \centering
    \includegraphics[width=\columnwidth]{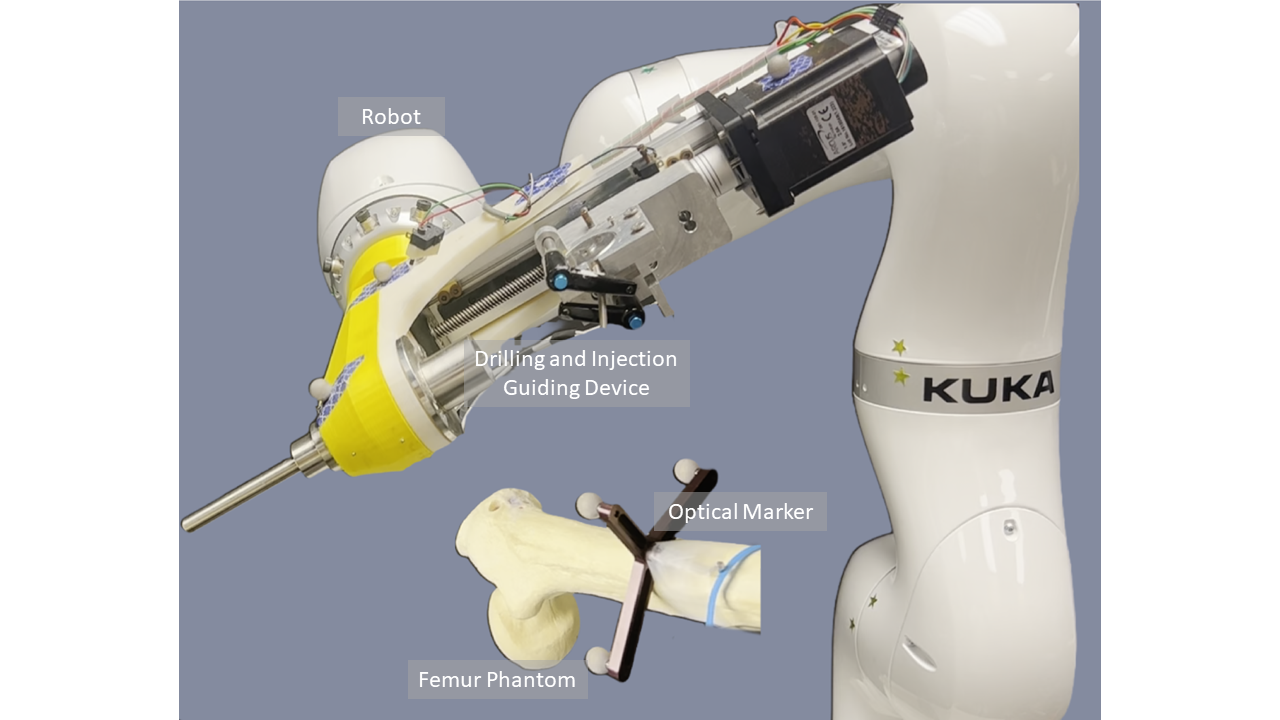}
    \caption{The robotic-assisted drilling and injection system \cite{bakhtiarinejad2023surgical}. Only the robot, drilling and injection guiding device, and the femur phantom are shown in this figure. The contents in the figure are integrated into the proposed PCMRS system. }
    \label{fig:FemurSetup}
\end{figure}

\begin{figure}
    \centering
    \includegraphics[width=\columnwidth]{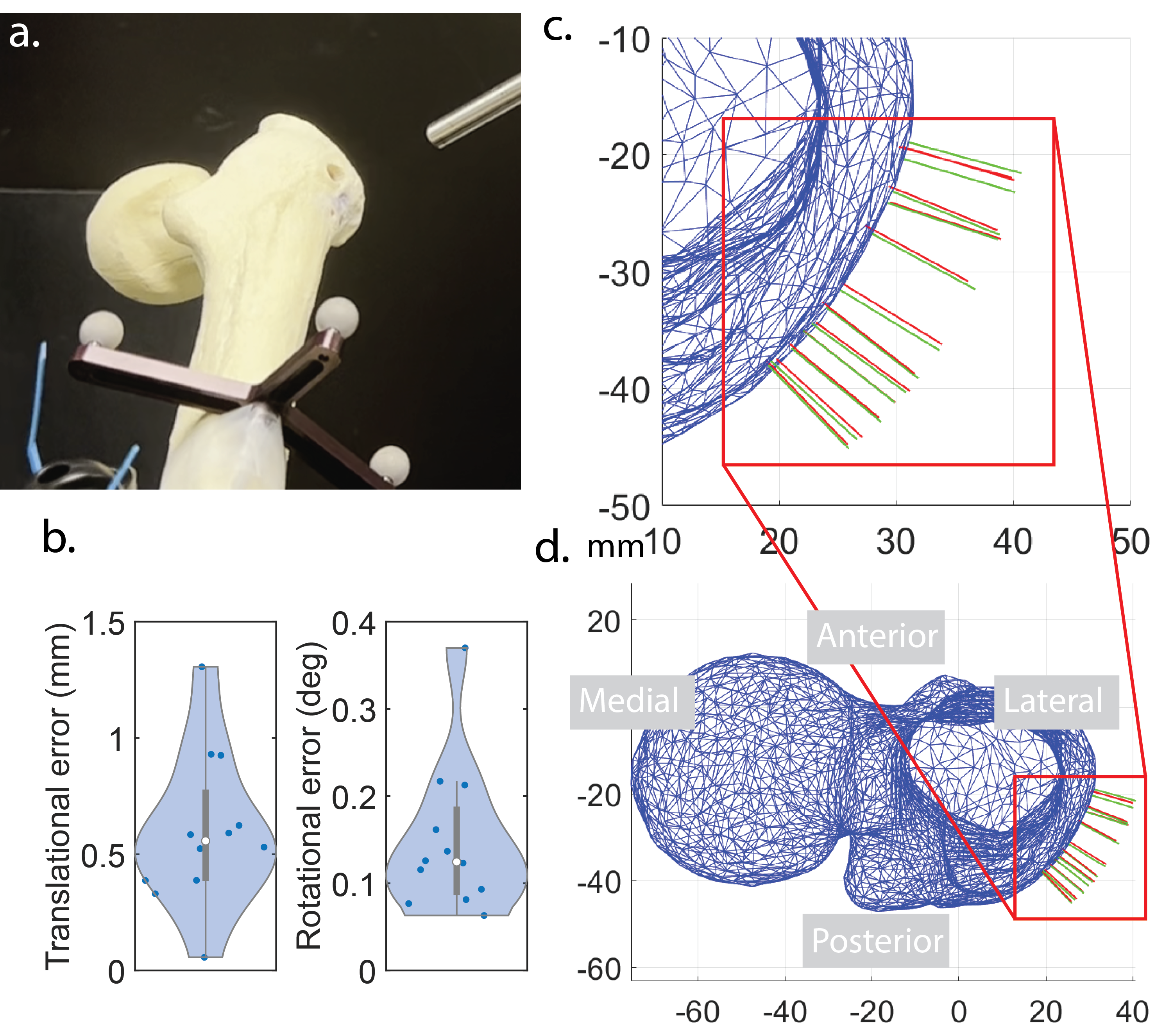}
    \caption{Placement errors in 12 trials for the femur phantom used to demonstrate the alignment task in osteoporotic hip augmentation. Panel a. shows the femur phantom and panel b. shows the errors between the planned tool poses and the measured tool poses. The translational errors are in Euclidean distances, and the rotational errors are angular errors in angle-axis representation. Panel c. and d. show the top view of the processed mesh model from the CT scan. The planned tool poses are shown in the green lines and the measured poses in the red lines.}
    \label{fig:femur}
\end{figure}

Similarly to Section \ref{sec:TMSexample}, here we mount the drill guide used in \cite{bakhtiarinejad2023surgical}, integrate the workflow, and test 12 alignments. The robotic setup is shown in Fig. \ref{fig:FemurSetup}. The average translational error (Euclidean) is 0.5979 mm, with a standard deviation 0.3280 mm, and the average rotational error (angle-axis representation) is 0.1480 degrees, with a standard deviation of 0.0855 degrees. Fig. \ref{fig:femur} shows the planned and measured poses and their relative errors.

Failure injection experiments are performed using the same method in Section \ref{sec:TMSexample}. In Table \ref{tab:failinj}, experiments \# 4-6 are the control group, corresponding to the experiments performed without injected errors. Experiments \#8, 12-14, and 18-20 correspond to failure injection experiments. Injected error types are the same to the ones in Section \ref{sec:TMSexample}. The results show that the system successfully performs the workflow and can reject injected errors. 

\subsection{Visualization and Communication}

The 3D Slicer bridging module \textit{simple-slicer-comm} transmits data to the ROS dispatcher module. The data include the planned landmarks coordinates, the planned target tool poses, and user commands such as performing registration and starting visualization. With the addition of \textit{simple-ros-comm}, the visualization of the current and the target tool poses can be done in real-time. The frame rate is stable at least 60 Hz in both 3D Slicer and Unity Editor simulation, shown in Fig. \ref{fig:UI}.

PCMRS uses UDP for its communication modules since the use cases of PCMRS are mostly robotic applications, where sensors are streamed at a high frequency, and packet latency is more important than reliable transport. ROS adopts UDP in its transport layer for ROS topics and services due to the same reason. However, users can implement TCP instead of UDP in PCMRS if the need is to focus on the reliable delivery of data packets. In future versions of PCMRS, an additional TCP transmission option can help with delivery-critical tasks. Any modules connected by UDP or TCP in PCMRS can be hosted on different machines. A demonstration of the connection between \textit{dispatcher} and \textit{Unity} is shown in Fig. \ref{fig:UI}. Network Address Translation Traversal (NAT-T) services allow geographically separated machines to be on the same local area network, so teleoperation is possible for modules using TCP or UDP. Other modules connected by ROS topics in PCMRS can use the networking functions in ROS to realize teleoperation.

\section{Conclusions}

This article applies controlled processes in medical robotics, introduces the procedures of designing a process-controlled medical robotics system, describes the philosophy for implementation, and provides the software design in detail. With the standardization in our work, automation with process control is possible, where fully deterministic steps (from hFSM to machine executables) and semi-deterministic (clinician experience to hFSM) have been identified. The proposed system has the integrated capabilities of planning, perception, actuation, visualization, and teleoperation. The hFSM design and D-FSO paradigm are easy to re-purpose and extend, and we have demonstrated two medical applications using the design. During the experiments, PCMRS executes procedures strictly following the state transitions defined by their corresponding hFSMs and prevents invalid operations. Additionally, the accuracy of the robot, integrated with sensors and kinematic calculations have been tested.  

The hierarchical FSM is not limited to software design but has profound implications for automating medical procedures. From the perspective of automation, constructing a clear FSM can translate human experience to machine-executable procedures. The translation is deterministic and can be automated. With the design cycle discussed in Section \ref{sec:methods}, this work contributes to the possibilities of a higher level of automation in medical settings.

\bibliographystyle{IEEEtran}
\bibliography{bib} 

\newpage

\begin{IEEEbiography}[{\includegraphics[width=1in,height=1.25in,clip,keepaspectratio]{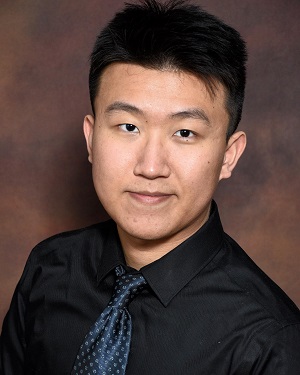}}]{Yihao Liu} is a Ph.D. student in the Department of Computer Science in the Johns Hopkins University. He received his M.Sc. degree in robotics from the Johns Hopkins University in 2021 and B.ASc degree in Electrical Engineering, minor in Computer Science from the University of British Columbia in 2019. His research interests include medical/surgical robotics, augmented reality and other engineering applications in health care.
\end{IEEEbiography}

\begin{IEEEbiography}[{\includegraphics[width=1in,height=1.25in,clip,keepaspectratio]{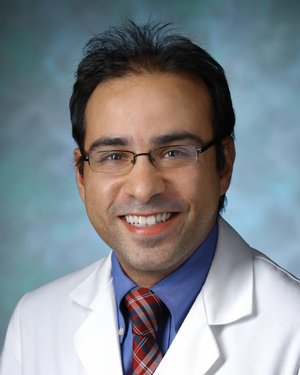}}]{Amir Kheradmand, M.D.} is an associate professor of neurology, neuroscience, and otolaryngology-head \& neck surgery in School of Medicine, Johns Hopkins University. He is the director of research in the neuro-visual \& vestibular division. His research and interests include perception of spatial orientation, neural mechanism of orientation constancy, functional brain mapping of upright perception, neural mechanism of spatial disorientation in patients with dizziness, and cular motor control and corollary visual-vestibular sensory processing.
\end{IEEEbiography}

\begin{IEEEbiography}[{\includegraphics[width=1in,height=1.25in,clip,keepaspectratio]{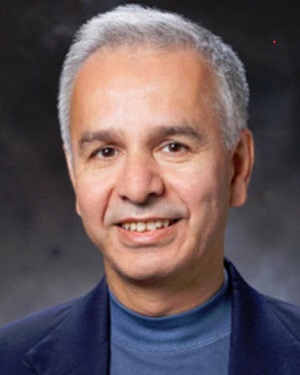}}]{Mehran Armand, Ph.D.} received the Ph.D. degree in mechanical engineering and kinesiology from the University of Waterloo, Waterloo, ON, Canada, in 1998. He is currently a Professor of Orthopaedic Surgery, Mechanical Engineering, and Computer Science with Johns Hopkins University and a Principal Scientist with the JHU Applied Physics Laboratory. Prior to joining APL in 2000, he completed postdoctoral fellowships with the JHU Orthopaedic Surgery and Otolaryngology-Head and Neck Surgery. He currently directs the Laboratory for Biomechanical- and Image-Guided Surgical Systems, JHU Whiting School of Engineering. He also directs the AVICENNA Laboratory for advancing surgical technologies, Johns Hopkins Bayview Medical Center. His laboratory encompasses research in continuum manipulators, biomechanics, medical image analysis, and augmented reality for translation to clinical applications of integrated surgical systems in the areas of orthopaedic, ENT, and craniofacial reconstructive surgery.

\end{IEEEbiography}

\vfill

\end{document}